\documentclass[conference]{IEEEtran}
\IEEEoverridecommandlockouts
\usepackage{ragged2e}
\usepackage{listings}
\usepackage{float}
\usepackage{cite}
\usepackage{amsmath,amssymb,amsfonts}
\usepackage{algorithmic}
\usepackage{graphicx}
\usepackage{textcomp}
\usepackage{xcolor}
\def\BibTeX{{\rm B\kern-.05em{\sc i\kern-.025em b}\kern-.08em
    T\kern-.1667em\lower.7ex\hbox{E}\kern-.125emX}}

\makeatletter % changes the catcode of @ to 11
\newcommand{\linebreakand}{%
  \end{@IEEEauthorhalign}
  \hfill\mbox{}\par
  \mbox{}\hfill\begin{@IEEEauthorhalign}
}
\makeatother % changes the catcode of @ back to 12
    
\begin{document}

\title{Botson: An Accessible and Low-Cost Platform for Social Robotics Research\\
}

\author{
\IEEEauthorblockN{Samuel Bellaire\IEEEauthorrefmark{1}}
\IEEEauthorblockA{srbellai@umich.edu}

\and

\IEEEauthorblockN{Abdalmalek Abu-raddaha\IEEEauthorrefmark{1}}
\IEEEauthorblockA{abdmalek@umich.edu}

\and

\IEEEauthorblockN{Natalie Kim\IEEEauthorrefmark{1}}
\IEEEauthorblockA{nbkim@umich.edu}

\linebreakand

\IEEEauthorblockN{Nathan Morhan\IEEEauthorrefmark{1}}
\IEEEauthorblockA{nmorh@umich.edu}

\and

\IEEEauthorblockN{William Elliott\IEEEauthorrefmark{1}}
\IEEEauthorblockA{welliot@umich.edu}

\and

\IEEEauthorblockN{Samir Rawashdeh\IEEEauthorrefmark{1}}
\IEEEauthorblockA{srawa@umich.edu}

\linebreakand

\IEEEauthorblockN{\IEEEauthorrefmark{1}University of Michigan-Dearborn %\\ Dearborn, MI, USA
}
}

\maketitle

\begin{abstract}
Trust remains a critical barrier to the effective integration of Artificial Intelligence (AI) into human-centric domains. Disembodied agents, such as voice assistants, often fail to establish trust due to their inability to convey non-verbal social cues. This paper introduces the architecture of Botson: an anthropomorphic social robot powered by a large language model (LLM). Botson was created as a low-cost and accessible platform for social robotics research.
\end{abstract}


\section{Introduction}
\label{sec:intro}

The proliferation of disembodied artificial intelligence (AI), such as voice assistants and chatbots, has had a marked impact in human-robot interaction (HRI). However, their utility is often hampered by a "trust deficit," limiting their potential for collaborative partnerships. Trust, defined as a willingness to be vulnerable to another's actions under uncertainty, is fundamental for effective collaboration \cite{jacovi2021formalizing}. Existing works in the literature note that perceived transparency, explainability, and accuracy in AI systems are among the largest antecedents of trust in HRI, and deficiencies in any of these tends to erode trust significantly \cite{afroogh2024trust}. This degradation of trust often renders human-AI interactions transactional and impersonal. This leads to reduced reliance on AI, and correspondingly to under-utilization of AI systems \cite{lee2004trust}.

Another key aspect impacting trust in HRI is embodiment, or the lack thereof. Disembodied agents are constrained to verbal and textual modalities, stripping them of the rich, non-verbal channels --- such as gestures, gaze, and posture --- that are foundational to building rapport and trust in human social dynamics \cite{morency2010modeling}. Physical embodiment offers a solution to this challenge. By utilizing an AI with a physical form, the interaction is transformed from a disembodied command-response loop into a tangible social encounter\cite{correia2020dark}. Physical robots can leverage non-verbal cues to enhance communication, with studies consistently showing that users perceive embodied agents as more engaging and socially present than their disembodied counterparts. \cite{pinto2025designing}.

This paper introduces Botson: a social robot powered by a large language model (LLM), designed as an accessible and replicable platform to investigate the impacts of physical embodiment on human-AI trust. The system is built on a low-cost hardware stack, utilizing a Raspberry Pi, an Arduino micro-controller, and a 3D-printed humanoid chassis \cite{ottoDIY}. Its core mechanism generates emotionally congruent multimodal behavior: a user's spoken input is processed by the LLM, which is prompted to generate both a verbal response and a corresponding sentiment label (e.g. "happy," "serious"). This label triggers a synchronized physical gesture, leading to a more natural and believable interaction.

Our work makes two primary contributions to the field of HRI. First, we present the architecture of a low-cost, open-source system that integrates a cloud-based LLM with a physically embodied robot to enable real-time, sentiment-driven behavior. Second, we demonstrate an efficient prompt engineering technique that elicits both a verbal response and an emotional classification in a single inference step, offering a computationally lightweight method for affective robotics.

\section{Background \& Related Works}
\label{sec:background}

Our work is situated at the intersection of four key domains: the psychology of trust in HRI, the impact of robot embodiment and appearance, the development of accessible robotic research platforms, and the use of LLMs for affective robotics.

\subsection{Effects of Embodiment on Trust in HRI}
\label{sec:embodiment}

Trust in HRI is a complex and dynamic construct, ranging from general interpersonal models to application-specific frameworks. Foundational models typically define trust as a function of perceived ability and integrity \cite{svare2020function}. While the perception of a robot's ability is often determined by performance-based attributes such as explainability and accuracy, Kaplan et al. note that certain physical attributes, such as anthropomorphic --- or human-like --- embodiment, have a large impact on human-AI trust \cite{kaplan2023trust}. The authors also note that personality and behavior --- in which non-verbal cues play a significant role --- are strong contributors to trust as well.

This finding is reinforced in a large number of studies in the literature. Biswas et al. utilized the Pepper robot to play rock-paper-scissors with human participants, and found that users preferred to play the game with Pepper over a disembodied AI \cite{biswas2024robots}, citing increased engagement with and lower suspicion of cheating by the embodied AI agent. Babel et al. found that, in some HRI tasks, gaze control played a significant role in trust, with participants preferring more human-like gaze behaviors when engaging in social encounters with the robot \cite{babel2021small}. Bernotat et al. examined the effect of body shape on the willingness of participants to use a robot for specific tasks, and found that users were more inclined to use distinctly feminine robots for a wider array of tasks compared to more masculine robots \cite{bernotat2021fe}.

Although anthropomorphic robots are generally perceived more positively compared to their disembodied counterparts, studies have also shown that a balance must be achieved with other more "robotic" traits. In a trial, Złotowski et al. discovered that many users find highly human-like robots to be unsettling, and trust was overall lower compared to more stereotypical "robotic" appearances \cite{zlotowski2016appearance}. This is also noted by Barrow et al., who found that more "baby-like"  expressions with exaggerated facial features garnered more trust in users who interacted with the robot, in comparison with more realistic facial proportions \cite{barrow2025influence}.

This concept of developing anthropomorphic robots without straying into realistic human-like designs has influenced how we have developed the chassis for Botson, which is shown in Sec. \ref{sec:design}.

\subsection{Enabling Technologies for Affective Social Robots}
\label{sec:enabling}

The development of Botson is part of a broader movement toward democratizing robotics research through low-cost, open-source platforms. Cost-effective manufacturing technologies, such as 3D printing, allow roboticists to rapidly prototype and fabricate custom robot chassis \cite{saini2021role}. Low-cost single-board computers (SBCs) and micro-controllers, such as the Raspberry Pi and Arduino, have allowed for increased accessibility to social robotics, as well as a wider range of domains. This approach is seen in some existing robotic platforms, like Poppy \cite{lapeyre2014poppy} and ToddlerBot \cite{shi2025toddlerbot}, with both having key design constraints in the form of affordability and reproducibility.

Another key technology that has revolutionized social robots is the LLM. A key challenge in social robotics is the production of meaningful and believable multimodal behavior. Existing works in the literature show that LLMs are adept in generating emotionally congruent non-verbal cues, such as facial expressions or gestures \cite{kim2024understanding, pinto2025designing}. For example, Mishra et al. conducted a study using sentiments returned by a LLM to display emotions on a AI agent. They found that users preferred engaging with the emotive agent compared to a non-emotive version \cite{mishra2023real}. Recent studies have also found that LLMs are able to achieve co-speech gesture synthesis, as demonstrated in \textit{LLM Gesticulator} by Pang et al. \cite{pang2025llm}.

\section{System Design}
\label{sec:design}

\begin{figure}[htb]
    \centering
    \includegraphics[width=0.45\textwidth]{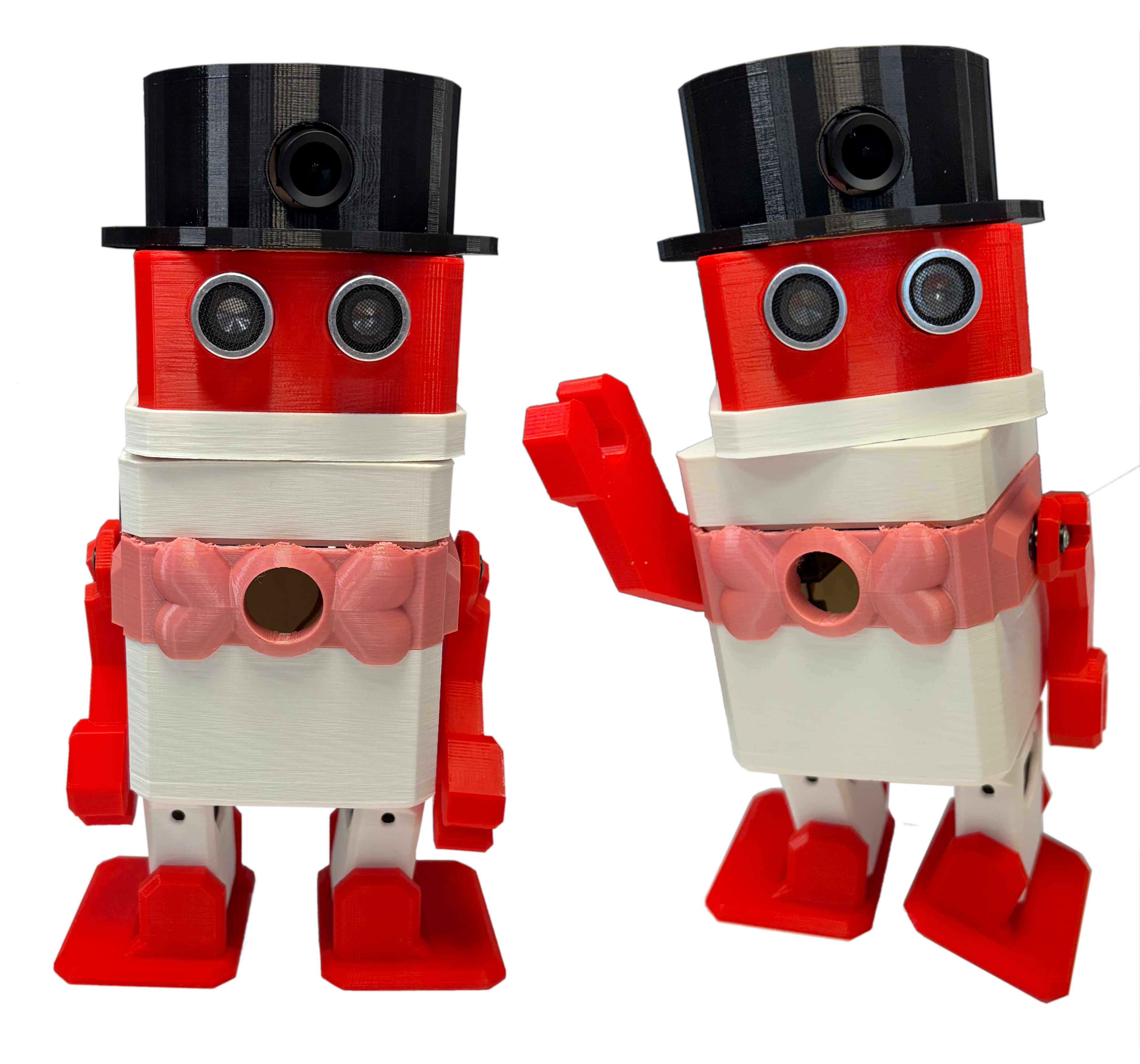}
    \caption{Botson in Idle State (left) and Performing a Waving Gesture (right)}
    \label{fig:botson}
\end{figure}

Botson's physical form is an anthropomorphic chassis based on a modified version of Otto DIY \cite{ottoDIY}, as shown in Fig. \ref{fig:botson}. The mechanical design consists of modular parts that can be attached in an arbitrary manner. The design is "robotic" in nature, as previous studies discussed in Sec. \ref{sec:embodiment} note that anthropomorphic designs with distinctly robotic features are often the most performant in regards to building trust in HRI. The chassis contains 7 motors responsible for actuating the robot's limbs. Four servo motors are used to control the robot's lower extremities --- one for each foot and an additional motor for each leg. Botson also employs two servo motors to move its arms, and a single motor to allow for neck rotation.

\begin{figure}[htb]
    \centering
    \includegraphics[width=0.48\textwidth]{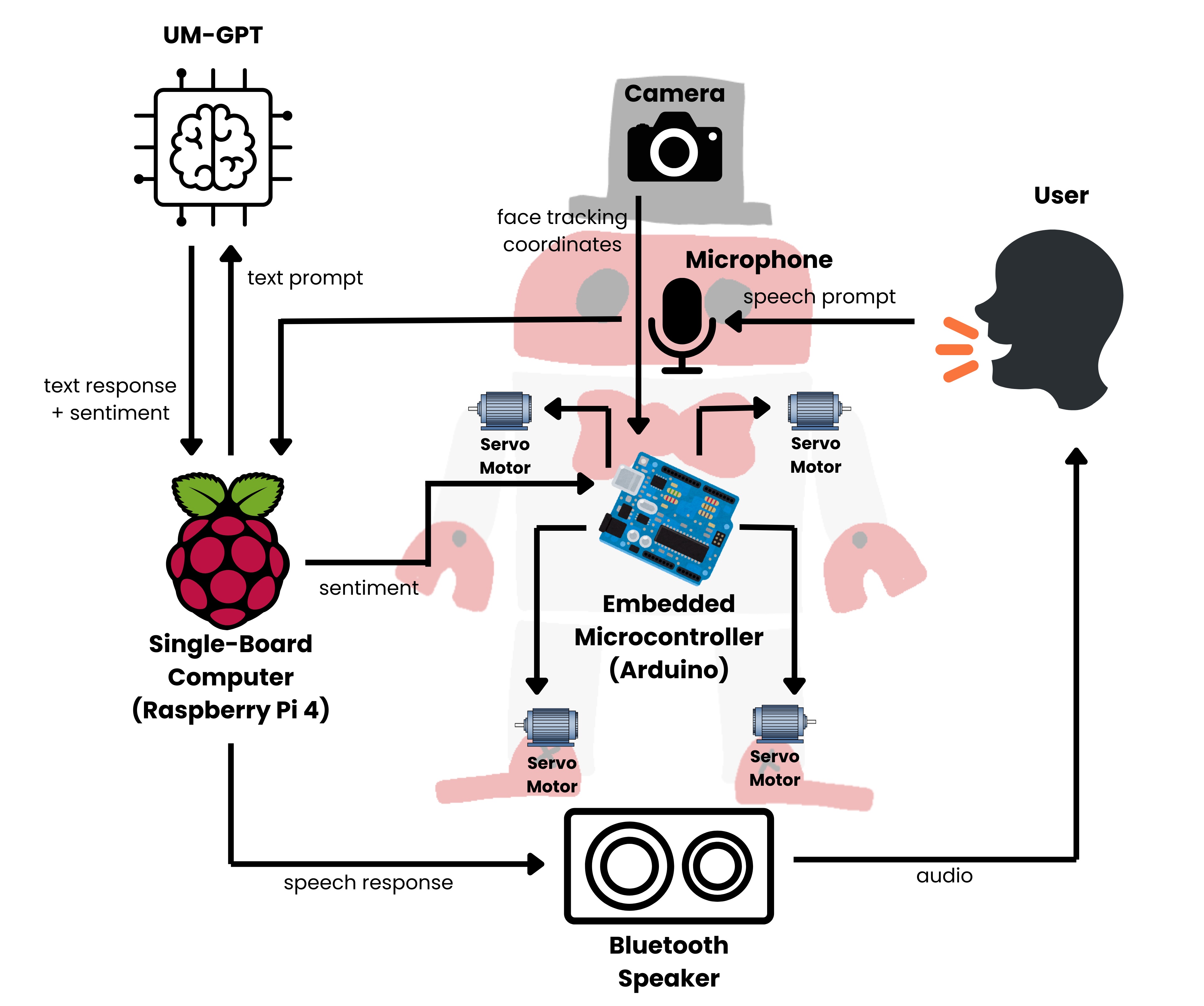}
    \caption{High-Level Block Diagram of Botson}
    \label{fig:block_diagram}
\end{figure}

The high-level overview of Botson's design is shown in Fig. \ref{fig:block_diagram}. An on-board Arduino micro-controller is responsible for the control of all seven actuators in the robot's chassis. The micro-controller is also connected to an offboard Raspberry Pi 4 SBC via a USB 2.0 connection. This allows Botson to offload more computationally-demanding tasks, such as the LLM pipeline, to an external platform. The SBC also handles all of the user inputs: a lavalier microphone and bluetooth speaker are the user's primary interface to the system's LLM pipeline, and a camera is also integrated into Botson's chassis to allow for additional future functionalities, such as head tracking or integration with a vision-language model (VLM).

\subsection{LLM Pipeline}
\label{sec:llm}

\begin{figure}[htb]
    \centering
    \includegraphics[width=0.48\textwidth]{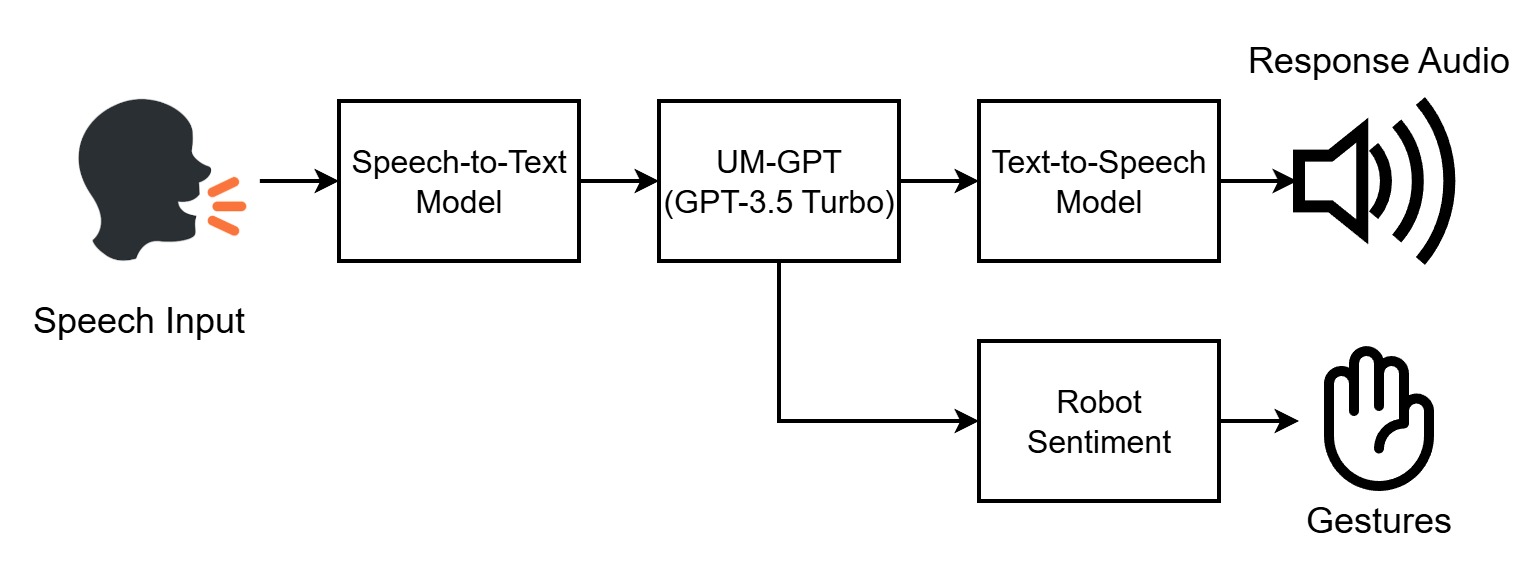}
    \caption{Flowchart of LLM Pipeline}
    \label{fig:llm_pipeline}
\end{figure}

Botson currently utilizes UM-GPT, a toolkit provided by the University of Michigan that offers access to a number of different AI services and models. Botson employs GPT-4o for its LLM pipeline, which is shown in Fig. \ref{fig:llm_pipeline}.

Users initiate an interaction with Botson by pressing a push-button. Existing works in the literature examine the possibility for robots to initiate conversations themselves \cite{avelino2021break}, though Avelino et al. note that there are still significant challenges in proactive robots due to the complexity of social and environmental contexts that define human social dynamics. Thus, to ensure the intent of the user to interact with Botson, we use this button press to signify the beginning of a conversation.

After initiating, the user speaks to the robot and the speech is recorded by Botson's microphone until the user is finished speaking and remains silent for at least one second. Afterwards, the \texttt{speech\_recognition} library in Python is used to make a call to Google's speech-to-text API. Once the user's speech is recorded and transcribed, the text is sent to the LLM along with the conversation history. For new interactions with the robot, the history consists solely of the pre-prompt given to the LLM, which is outlined in Sec. \ref{sec:prompt}. As the conversation unfolds, the dialogue is continually appended to the history.

The LLM returns its response as a JSON object containing both the LLM's text response, as well as a sentiment picked from a pre-defined list of 5 emotions. The text response is passed to an eSpeak text-to-speech process that is configured to use a distinctly robotic voice, while the sentiment value is sent to the Arduino micro-controller for further handling, which is detailed in Sec. \ref{sec:gesture}.

\subsection{Prompting Framework}
\label{sec:prompt}

Botson utilizes a lightweight prompting framework as part of its design. In its initial context, the LLM is informed of its name and the role that it is to fulfill. In our case, we inform the LLM that it is a helpful assistant embedded into an embodied robot. We also indicate that we use a speech-to-text model at the input of the pipeline, and instruct it to compensate for potential errors in the transcription process. At the end of the pre-prompt, we instruct the model to generate and return a JSON object containing both its textual response, as well as a sentiment chosen from a list of 5 different emotions, outlined in Sec. \ref{sec:gesture}.

We also take measures to limit the length of Botson's outputs. Response length is a challenge that many social robots face, as LLMs tend to favor returning a high volume of information in response to certain prompts \cite{poddar2025brevity}. These long-winded responses are undesirable for social robots, since they often lead to one-sided conversations that reduce the amount of meaningful dialogue a user can have with the robot. With Botson, we circumvent this challenge in the pre-prompt. As part of the robot's initial context, we instruct Botson to keep all of its responses to within approximately 30 words, and specifically ask the LLM to carry its discussion over multiple responses in a conversational style. By doing this, the LLM is able to cut down its overly-verbose responses into just one or two sentences, facilitating a more realistic and believable conversation. The full initial context provided to the LLM is given below.

\textit{\small
Your name is Botson. You are the brains of a helpful assistant embodied into a robot. We are using speech-to-text, so there may be some errors. Do your best to mitigate, or ask for clarification. Keep your responses within 30 words. When you have more to say, share it over multiple responses in a conversational style. The response should include one text response. Each response must end with a single character that is most appropriate from this list: a (greeting), b (happy), c (sad), d (serious), e (dance). When responding, use the following JSON format, where x is the response, and y is the sentiment from the provided list: \{ "Response": "x", "Sentiment": "y"\}
}

\subsection{Gesture Synthesis}
\label{sec:gesture}

Alongside its audio output, Botson can also perform a number of distinct gestures based on the sentiment returned from the LLM. Currently, Botson supports 5 pre-defined gestures: greeting, happy, sad, serious, and dance. The joint angles for each of the 5 gestures are hard-coded in the Arduino micro-controller, though some randomization is applied to these movements  to produce a less rigid and more believable interaction with the user. In the LLM pipeline, the sentiment of the model's response is separated from the text, and sent to the Arduino micro-controller to be translated into the corresponding gesture.

\section{Pilot Study Results \& Analysis}
\label{sec:pilot}

After creating Botson, we conducted a preliminary usability study with 6 subjects to gage how they interact with the robot. Users were tasked with completing a simple 5x5 crossword puzzle, and were instructed to use both Botson, as well as a voice-only version of ChatGPT running GPT-4o, to help them solve the puzzle. The order in which users interacted with both agents was randomized; 3 subjects used Botson first and transitioned to ChatGPT after partially solving the crossword puzzle, and the other 3 subjects started with ChatGPT first. After either completing the puzzle, or spending approximately 5 minutes with both AI agents, each user was asked to complete a brief survey, with most survey questions formatted as a likert scale from 1-5. No personal information was collected during this process.

\subsection{AI Agent Helpfulness}
\label{sec:helpful}

\begin{figure}[htb]
    \centering
    \includegraphics[width=0.38\textwidth]{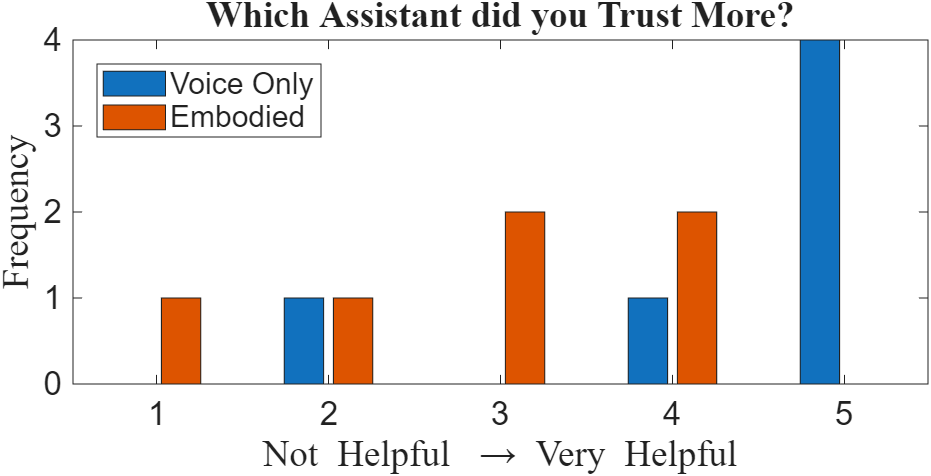}
    \caption{Evaluation of Helpfulness of AI Agents by Users}
    \label{fig:helpfulness}
\end{figure}

\begin{figure}[htb]
    \centering
    \includegraphics[width=0.38\textwidth]{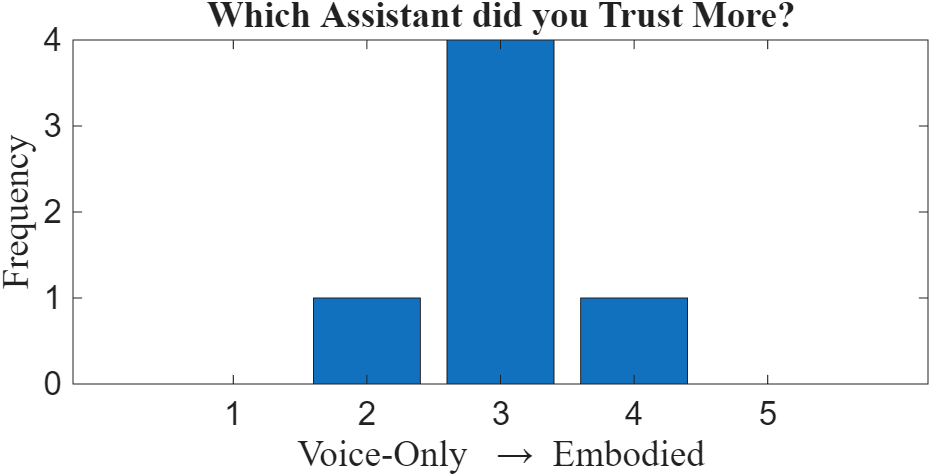}
    \caption{Evaluation of Trust in Voice-Only vs. Embodied AI}
    \label{fig:trust}
\end{figure}

In the first half of the survey, participants were questioned about the helpfulness and usefulness of both agents, as well as which one they found to be more helpful in their task. Participants found the voice-only agent to be more helpful overall (average score of 4.33) than the embodied robot agent (average score of 2.83), as shown in Fig. \ref{fig:helpfulness}. However, users did not give preference on average for either agent when asked which they trusted more to give correct or useful answers (average score of 3.00, where 1 indicates strong preference for the voice-only agent, 3 is neutral, and 5 indicates strong preference towards Botson). This is shown in Fig. \ref{fig:trust}.

Since both agents employed GPT-4o, we expect that the quality of their answers should be consistent due to utilizing the same model; this coincides with the neutral perception on which agent gave more useful answers. In terms of the overall helpfulness, we observed that users preferred the voice-only agent due to some characteristics of the embodied robot. In an open-ended question at the end of the survey, participants found the conversation with ChatGPT to flow better, especially due to its more natural sounding voice compared to Botson.

\subsection{Embodied Robot Engagement}
\label{sec:engagement}

\begin{figure}[htb]
    \centering
    \includegraphics[width=0.38\textwidth]{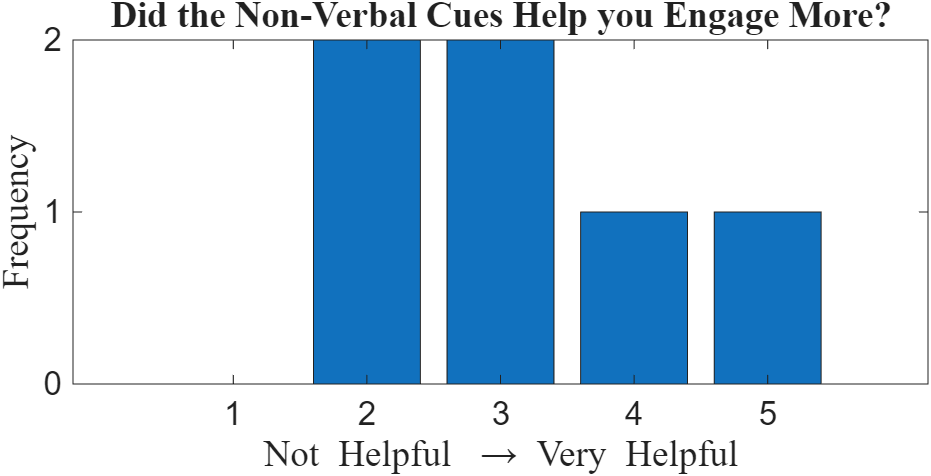}
    \caption{Usefulness of Gestures in Engagement}
    \label{fig:nonverbal}
\end{figure}

\begin{figure}[htb]
    \centering
    \includegraphics[width=0.38\textwidth]{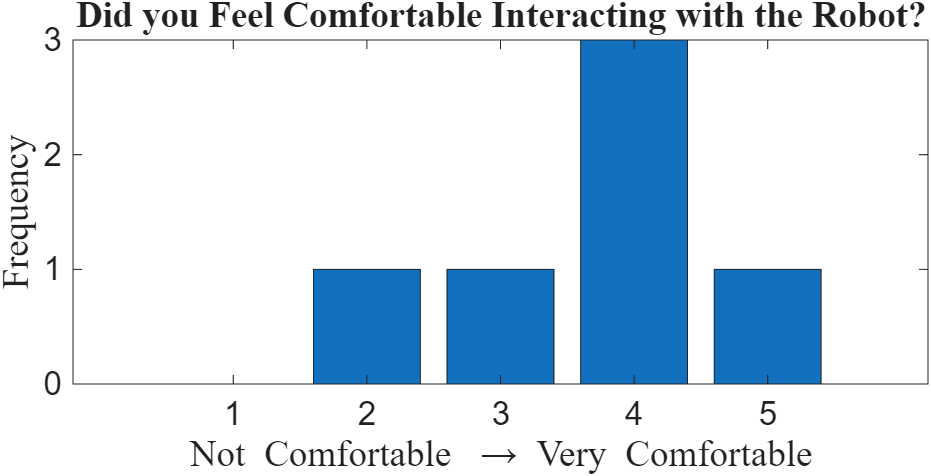}
    \caption{Comfort Level of Users with Embodied Robot}
    \label{fig:comfort}
\end{figure}

Users generally found that they were able to engage more with the embodied robot, with half of all participants indicating that they found the embodied robot to be more fun to interact with. Users also noted that the non-verbal gestures allowed them to engage more with the embodied robot (average score of 3.17), as shown in Fig. \ref{fig:nonverbal}. Participants also indicated that they were generally comfortable while interacting with the robot (average score of 3.67) in Fig. \ref{fig:comfort}.

In the open-ended question at the end of the survey, participants noted that they generally liked the gestures that Botson made, and that embodiment was somewhat more fun and engaging to interact with. However, one user indicated that the meaning of the robot's gestures was not clear in all cases.

\section{Conclusion \& Future Work}
\label{sec:conclusion}

In this paper, we presented the overall design and architecture of Botson: an anthropomorphic social robot integrating an LLM to create a low-cost and reproducible research platform for social robotics. As the project matures, we also plan to release the full, open-source design of Botson for other roboticists to utilize in their research work. This will be done as part of a future work that is currently underway.

In the near future, we plan to make further modifications and improvements to Botson. In particular, we aim to look at upgrading Botson's LLM pipeline to a more modern model, such as GPT-5.2. We also plan to switch to OpenAI's realtime API to address some of the user feedback we received in Sec. \ref{sec:pilot} to facilitate smoother conversation between the user and the embodied agent. In particular, users preferred the more natural-sounding voice of the voice-only agent as opposed to Botson's more robotic voice. Participants also preferred to have a seamless conversation with the voice-only agent instead of initiating with a button press, even though the voice-only agent sometimes interrupted them during a pause in speech.

Lastly, we plan to integrate a camera into the AI pipeline to give the system visual contexts to its conversations. After completing these, we plan to conduct a more thorough user study with more participants to gage the impact of embodiment, gestures, and emotions on how users perceive AI agents.

In addition to our plans to iterate upon certain functionalities of Botson, we also intend to target two specific domains of HRI research, where we will deploy Botson in user studies for development of novel technologies in social robotics. In Sec. \ref{sec:enabling}, we outline the LLM as one of the key enabling technologies of modern social robots, as they are adept at generating emotionally congruent multimodal behavior in embodied agents. We believe that for social robotics, gesture and expression synthesis are promising areas of future research. We discuss these briefly in Sec. \ref{sec:gesture_synth} and Sec. \ref{sec:expression_gen}.

\subsection{LLM-Based Gesture Synthesis}
\label{sec:gesture_synth}

In recent years, LLMs have gained traction in motion and gesture generation literature, where they are used in co-speech gesture synthesis \cite{pang2025llm}, as well as in the generation of motions from textual descriptions \cite{zhang2024motiongpt}. Although the concept of using LLMs as motion and gesture generators has seen significant advancement, the applications of many studies are limited to virtual agents. We believe that physically embodied robots offer a new frontier of novel, high-impact research in gesture synthesis applications. Botson can be useful in this domain as a low-cost and rapidly deployable platform for initial proof-of-concept investigations before more time is spent sourcing a more complex and costly robotic research platform.

\subsection{Dynamic Expression Generation}
\label{sec:expression_gen}

Alongside gestures, emotions are also important non-verbal cues in human social dynamics. Recent literature has examined the use of both foundational models \cite{chae2025affective}, as well as the integration of LLMs with other types of models \cite{zhao2023chatanything}, to generate human-like faces and emotions for various applications. We believe that real-time generation of facial expressions for physically embodied robots is a key step in developing social robots that are more engaging and trustworthy. Because of Botson's modular architecture, it allows for low-cost integration of displays on its chassis to display facial expressions, enabling research on expression synthesis for social robots.

\bibliographystyle{IEEEtran}
\bibliography{refs.bib}

\end{document}